%% file: acl_latex.tex
\documentclass[11pt]{article}

\usepackage[preprint]{acl}

\usepackage{times}
\usepackage{latexsym}

\usepackage[T1]{fontenc}

\usepackage[utf8]{inputenc}

\usepackage{microtype}

\usepackage{inconsolata}

\usepackage{graphicx}
\usepackage{xcolor}
\usepackage{subcaption}

\usepackage{amssymb}
\usepackage{amsmath}
\usepackage{mathtools}

\usepackage{enumitem}
\setlist{nolistsep}

\usepackage{algorithm}
\usepackage{algorithmic}

\usepackage{url}
\usepackage{hyperref}

\usepackage{tabularx}
\usepackage{booktabs}
\usepackage{multirow}

\newcommand{\secref}[1]{\S\ref{#1}}

\title{Where Does Authorship Signal Emerge\\in Encoder-Based Language Models?}

\author{Francis Kulumba \\
  Inria Paris \\
  Sorbonne Université \\
  \texttt{francis.kulumba@inria.fr} \\\And
  Guillaume Vimont \\
  IRIF \\\AND
  Laurent Romary \\
  Inria Paris \\\And
  Florian Cafiero \\
  LRE, EPITA \\
  Ecole nationale des chartes -- PSL}

\begin{document}
\maketitle
\begin{abstract}
    Authorship attribution models fine-tuned with the same pretrained encoder, data, and loss can differ four-fold in performance depending only on their scoring mechanism. We use mechanistic interpretability tools to explain this gap. Stylistic features such as word length, punctuation density, and function-word frequency are similarly available at every layer in every model we probe, including an off-the-shelf control encoder, suggesting that the gap is not explained by their linear readability. Instead, causal intervention shows that the scorer appears to determine where the encoder consolidates authorship signal. Mean pooling forces consolidation by early to mid layers, while late interaction defers it to later layers. We further derive this difference from the gradient structure of each scorer, and training dynamics reveal distinct learning trajectories that follow from that difference.
\end{abstract}

\input{latex/1_introduction}

\input{latex/2_background}

\input{latex/3_theory}

\input{latex/4_setup}

\input{latex/5_results}

\input{latex/6_related_work}

\input{latex/7_discussion}

\section*{Limitations}

\paragraph{Backbone choice.}
We fix the backbone to ModernBERT to control for architecture, since our goal is to isolate how the scoring mechanism shapes signal consolidation. The specific inflection layers we observe, such as layer 9 for pooling and layers 15 to 16 for interaction, may shift in other architectures. The qualitative gap between early consolidation under mean pooling and later consolidation under interaction should however, transfer. Testing a second backbone, such as RoBERTa~\citep{liu_roberta_2019}, would be useful for architectural generality, but it is orthogonal to the main question of this paper.

\paragraph{Patch-level interaction.}
We study only $n=2$ for PLI. This keeps the analysis focused on the contrast between pooling and interaction while still giving us a middle regime to compare against LI and mean pooling. The theory suggests that larger patches should move the inflection earlier, closer to the pooling regime. Exploring $n=3,4,5$ would be a natural extension, but it is not necessary for the main result reported here.

\paragraph{Probe set size.}
The 148 triplets are enough to resolve the six-layer gap, but they are too small for fine-grained LI versus PLI comparisons. Bootstrap confidence intervals may not separate a one to two layer difference cleanly. The high failure rate on Tier B also leaves only 28 to 33 correctly ranked triplets, which makes those curves noisier than Tiers A and C. The main qualitative result is stable across all three tiers, but finer distinctions between LI and PLI remain below our statistical resolution. Considering these results, the probe results should be read as evidence about linear accessibility of stylistic features studied in this paper, not as a claim that the full hidden representations are equivalent across model families.

\section*{Acknowledgments}

The authors are grateful to Djamé Seddah who indirectly inspired this work. We also thank Wissam Antoun, Rian Touchent and Théo Lasnier for the productive discussions. This work was partially realized on computing HPC and storage resources provided by IDRIS thanks to the grant GCDA1016807 on the DALIA supercomputer.

\bibliography{custom}

\appendix

\section{Top LISA features across models}
\label{sec:app:top-lisa}

\begin{table}[h]
\centering
\small
\setlength{\tabcolsep}{4pt}
\renewcommand{\arraystretch}{1.15}

\begin{tabular}{clp{0.55\columnwidth}}
\toprule
\textbf{Rank} & \textbf{Feature} & \textbf{Peak $R^2$ by model} \\
\midrule

1 & wl\_mean &
\shortstack[l]{
Layerwise: .575 \\
LI: .580 \\
PLI $n{=}2$: .579 \\
E5: .576
} \\

\addlinespace[0.5em]

2 & fw\_we &
\shortstack[l]{
Layerwise: .499 \\
LI: .500 \\
PLI $n{=}2$: .505 \\
E5: .493
} \\

\addlinespace[0.5em]

3 & fw\_the &
\shortstack[l]{
Layerwise: .496 \\
LI: .496 \\
PLI $n{=}2$: .496 \\
E5: .496
} \\

\addlinespace[0.5em]

4 & punct\_period &
\shortstack[l]{
Layerwise: .493 \\
LI: .468 \\
PLI $n{=}2$: .502 \\
E5: .479
} \\

\addlinespace[0.5em]

5 & punct\_colon &
\shortstack[l]{
Layerwise: .473 \\
LI: .473 \\
PLI $n{=}2$: .473 \\
E5: .463
} \\

\bottomrule
\end{tabular}

\caption{Top-5 LISA features by peak $R^2$ across layers. All four models surface the same feature family with highly similar probe performance.}
\label{tab:lisa-top5}
\end{table}

Table~\ref{tab:lisa-top5} reports the top-5 LISA features by peak $R^2$ for each model. The rankings are nearly identical: mean word length dominates in all four models ($R^2 \approx 0.576$--$0.580$), followed by function-word frequencies and punctuation density. The control E5 encoder, which has never been trained on authorship data, achieves the same $R^2$ values as the three fine-tuned models, confirming that these stylistic features are linearly readable from the pretrained backbone and are not created via fine-tuning.

\end{document}

%% file: latex/1_introduction.tex
\section{Introduction}
\label{sec:introduction}

\begin{figure*}[t]
\centering
\includegraphics[width=\textwidth]{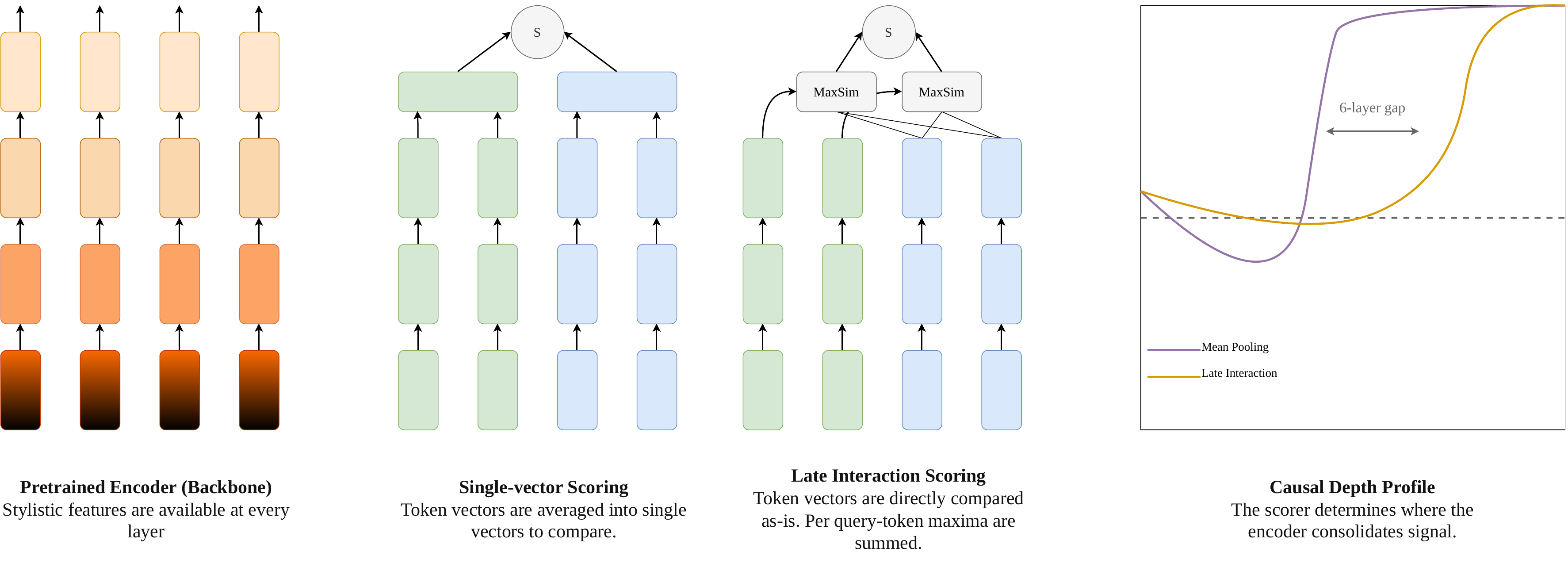}
\caption{\textbf{Conceptual overview.} \textbf{Left}: The pretrained language model encodes stylistic features at every layer, regardless of fine-tuning. \textbf{Center}: Two scoring mechanisms read out these features differently. Mean pooling averages all tokens into a single vector. Late interaction (LI)~\citep{khattab_colbert_2020} compares tokens directly. \textbf{Right}: Causal intervention reveals that the scoring mechanism appears to determine where the encoder consolidates authorship signal. Mean pooling forces early consolidation while $\mathrm{MaxSim}$ allows for late consolidation.}
\label{fig:teaser}
\end{figure*}

Every author leaves traces in their writing. Sentence length, punctuation habits, function-word preferences, and word-length distributions all carry information about who wrote a text, even when two authors write about the same topic~\citep{mosteller_inference_1963, burrows_delta_2002, keselj_ngram_2003}. Authorship attribution (AA) is the task of deciding, given two passages, whether they were written by the same person or group. A useful task for forensic linguistics~\citep{dauber_git_2019} or historical document analysis~\citep{cafiero_why_2019} among other applications.
 
Modern AA systems follow a contrastive learning paradigm: a pretrained text encoder produces a representation for each passage~\citep{Vaswani_attention_2017, devlin_bert_2019}, and a scoring function compares the representations to produce a similarity score~\citep{wegmann_same_2022, ai_whodunit_2022, huertas-tato_isolating_2024, kantharuban_idiolex_2026}. The encoder is fine-tuned so that same-author passages score high and different-author passages score low. This setup works well, but recent work has revealed a striking puzzle about the scoring function. \citet{kulumba_halvest_2025} trained multiple models on a scholarly corpus in which topic is decorrelated from authorship, and found that the choice of scoring mechanism alone explains much of the observed four-fold performance gap. All the models share the same pretrained backbone, the same training data, and the same contrastive loss. The only difference is the pooling/scoring mechanism: one family of models averages all token representations into a single vector before scoring (mean pooling), while another compares token representations directly via late interaction (LI)~\citep{khattab_colbert_2020}.

Why does such a large gap emerge from what is, in principle, only a difference in the final comparison step? There are at least two plausible explanations. The first is that different scoring mechanisms cause the encoder to learn different internal representations during fine-tuning: mean pooling forces the encoder to discard fine-grained stylistic information that LI preserves. The second is that the encoder learns similar representations regardless of the scorer, and the gap arises purely from how those representations are read out at inference time. This paper uses the interpretability toolkit~\citep{alain_understanding_2017, vig_investigating_2020, belinkov_probing_2022, goldowsky-dill_localizing_2023, zhang_towards_2023} on the fine-tuned encoders from \citet{kulumba_halvest_2025} to distinguish between these two explanations. This allows us to test a dissociation between feature \emph{availability} and feature \emph{use} (Figure~\ref{fig:teaser}):
 
\begin{itemize}
    \item \textbf{Availability is invariant}, the same stylistic features (word length, capitalization, punctuation density, etc.) are linearly readable from the hidden states of all models at all layers, including a control encoder picked off the shelf. The pretrained backbone already encodes these features. Contrastive fine-tuning does not create them.
    
    \item \textbf{Use depends on the scoring mechanism}, as it determines where in the encoder authorship signal becomes causally necessary. Mean pooling consolidates authorship signal by mid layers, while LI defers consolidation to late ones. This gap can be explained by the gradient structure of the scoring functions.
\end{itemize}

Our results show that the choice of scoring function determines the effective depth of the encoder, the information the model can exploit, and the trajectory it follows during training. Understanding this mechanism clarifies why LI-based systems consistently outperform pooled representations in AA, despite relying on the same pretrained backbone.

%% file: latex/2_background.tex
\section{Background}
\label{sec:background}
 
This section defines the building blocks of the contrastive AA pipeline and the analysis tools we use to study it.
 
\subsection{Contrastive authorship attribution}
\label{sec:bg-contrastive}
 
In the contrastive formulation, training data consists of triplets $(a, p, n)$: an anchor passage $a$, a same-author positive $p$, and a different-author negative $n$. The encoder $f_\theta$ maps each passage to a sequence of token-level representations. A scoring function $s$ then compares the anchor's representation to the positive's and to the negative's, producing scalar similarity scores. Training minimizes the InfoNCE loss~\citep{oord_representation_2019}:
\begin{equation}
\footnotesize
\mathcal{L} = -\log \frac{\exp\bigl(s(a, p) / \tau\bigr)}{\exp\bigl(s(a, p) / \tau\bigr) + \displaystyle\sum_{n' \in \mathcal{N}} \exp\bigl(s(a, n') / \tau\bigr)}
\label{eq:infonce}
\end{equation}
where $\tau$ is a temperature parameter and $\mathcal{N}$ is the set of in-batch negatives: every non-positive passage in the batch serves as a negative. This loss pushes the anchor closer to the positive and farther from all negatives in the scoring space.

\subsection{Scoring mechanisms}
\label{sec:bg-scoring}
 
The encoder produces a sequence of token representations $\mathbf{H}^a = [\mathbf{h}_1^a, \ldots, \mathbf{h}_m^a] \in \mathbb{R}^{m \times d}$ for a passage of $m$ tokens with hidden dimension $d$. The scoring function determines how this matrix is turned into a scalar similarity. We study three families.
 
\paragraph{Mean pooling with cosine similarity.}
The passage representation is the mean of its token embeddings and the score is the cosine similarity between mean vectors. Mean pooling is the standard AA baseline~\citep{rivera-soto_learning_2021, wegmann_same_2022, kantharuban_idiolex_2026}. It compresses the entire token sequence into a single $d$-dimensional vector before scoring.
 
\paragraph{Late interaction ($\mathrm{MaxSim}$).}
The passage is represented by its full sequence of token embeddings, and the score is the sum over anchor tokens of the maximum cosine similarity to any candidate token~\citep{khattab_colbert_2020}:
\begin{equation}
s_{\text{LI}}(a, p) = \sum_{i=1}^{m_a} \max_{j \in [m_p]} \cos(\mathbf{h}_i^a, \mathbf{h}_j^p)
\label{eq:maxsim}
\end{equation}
Unlike mean pooling, LI preserves per-token structure through the scoring function: the encoder does not need to compress all the information.
 
\paragraph{Patch-level late interaction (PLI).}
A middle ground. The token sequence is partitioned into contiguous patches of size $n$. Each patch is mean-pooled, and $\mathrm{MaxSim}$ is applied at the patch level:
\begin{equation}
s_{\text{PLI}}(a, p) = \sum_{i=1}^{P_a} \max_{j \in [P_p]} \cos(\mathbf{p}_i^a, \mathbf{p}_j^p)
\label{eq:pli}
\end{equation}
where $\mathbf{p}_i = \frac{1}{n}\sum_{t \in \text{patch}_i} \mathbf{h}_t$ is the mean of the tokens within patch $i$. We use $n{=}2$ (bigram patches) in this study.
 
\subsection{Alignment and uniformity}
\label{sec:bg-au}
 
We use the alignment–uniformity framework of \citet{wang_understanding_2020}, where alignment $\alpha$ measures closeness of same-author pairs and uniformity $u$ measures how evenly representations spread on the hypersphere (lower is better for both).

\subsection{Residual stream patching}
\label{sec:bg-patching}
 
Residual stream patching~\citep{vig_investigating_2020, meng_locating_2022} is a causal intervention that measures the contribution of each encoder layer to the model's output. If we corrupt the input of the encoder and then restore one layer's activations to their clean values, how much of the model's correct behavior is recovered?
 
Concretely, given a triplet $(a, p, n)$, we define three forward passes. A \emph{clean pass} encodes the positive $p$ normally, producing hidden states $\mathbf{h}^{(\ell)}_{\text{clean}}$ at each layer $\ell \in \{0, 1, \ldots, L\}$. A \emph{corrupt pass} encodes the negative $n$ normally, producing $\mathbf{h}^{(\ell)}_{\text{corrupt}}$. A \emph{patched pass} at layer $\ell$ encodes the negative, but at layer $\ell$ replaces the negative's hidden states with those from the positive. The patched hidden state then propagates through the remaining encoder layers to produce a patched score $s_{\text{patched}}^{(\ell)}$.

The clean score is $s_{\text{clean}} = s(a, p)$ and the corrupt score is $s_{\text{corrupt}} = s(a, n)$. If patching at layer $\ell$ recovers the clean score, it means layer $\ell$ carries the information needed for correct authorship scoring. If patching makes no difference, the information was not yet consolidated at that layer.
 
\subsection{Recovery metrics}
\label{sec:bg-recovery}
 
We quantify recovery with two metrics.
 
\paragraph{Percentage recovery} is a standard metric introduced by \citet{meng_locating_2022}:
\begin{equation}
\text{Recovery}^{(\ell)}(\%) = \frac{s_{\text{patched}}^{(\ell)} - s_{\text{corrupt}}}{s_{\text{clean}} - s_{\text{corrupt}}} \times 100
\label{eq:pct-recovery}
\end{equation}
A value of 0\% means no recovery while 100\% means full recovery. Values can go outside $[0, 100]$ in some particular cases. The problem with this metric is that the denominator $s_{\text{clean}} - s_{\text{corrupt}}$ can be very small, especially for scoring functions like PLI whose scores are more compressed. When the denominator is near zero, even tiny score changes produce enormous percentage values.
 
\paragraph{Rank recovery} avoids this problem by asking a binary question: after patching at layer $\ell$, does the model still rank the positive above the negative?
\begin{equation}
\scriptsize
r_{\text{rank}}^{(\ell)} = \frac{1}{|\mathcal{T}_+|} \sum_{t \in \mathcal{T}_+} \mathbf{1}\!\bigl[s_{\text{patched}}^{(\ell)}(a_t, p_t) > s_{\text{patched}}^{(\ell)}(a_t, n_t)\bigr]
\label{eq:rank-recovery}
\end{equation}
where $\mathcal{T}_+$ is the set of triplets the clean model ranks correctly. This gives a value in $[0, 1]$ with 0.5 being chance. We use rank recovery for all main-text figures and report percentage recovery in the appendix.
 
\subsection{LISA probes}
\label{sec:bg-probes}
 
To separate feature availability from feature use, we train linear probes~\citep{alain_understanding_2017, belinkov_probing_2022} at each encoder layer. The probes are regression models mapping the mean-pooled hidden state at layer $\ell$ to scalar stylistic features. We report the coefficient of determination $R^2$ on a held-out set. The feature targets are inspired by the LISA framework from \citet{kantharuban_idiolex_2026} and include nine categories: word length, capitalization rate, type--token ratio, punctuation density, function-word frequency, sentence length, hedging markers, citation density, and discourse connectives. A high $R^2$ at layer $\ell$ means the feature is linearly separable from the representation. This is a necessary but not sufficient condition for the model to actually use that feature for scoring

%% file: latex/3_theory.tex
\section{Gradient Structure and the Consolidation Bottleneck}
\label{sec:theory}
 
This section develops a theory of {what we expect to find, before any experiment is run. The theory starts from the gradient of the scoring function and derives a prediction about where in the encoder authorship signal should be consolidated.
 
\subsection{How the gradient distributes across tokens}
\label{sec:theory-gradient}
 
The end-to-end gradient of the InfoNCE loss with respect to a single token representation $\mathbf{h}_j^a$ factors into two parts:
\begin{equation}
\frac{\partial \mathcal{L}}{\partial \mathbf{h}_j^a} = \underbrace{\frac{\partial \mathcal{L}}{\partial s}}_{\text{InfoNCE term}} \cdot \underbrace{\frac{\partial s}{\partial \mathbf{h}_j^a}}_{\text{Scorer term}}
\label{eq:chain}
\end{equation}
The InfoNCE term concentrates gradient on hard negatives. This term is identical across scoring mechanisms: it depends on the values, not on how the scores were computed. The scorer term determines how that gradient distributes across individual tokens, and this is where the three mechanisms diverge.
 
\paragraph{Mean pooling: dense, uniform gradient.}
Under mean pooling, the score depends on each token only through the mean. The partial derivative is:
\begin{equation}
\frac{\partial s_{\text{mean}}}{\partial \mathbf{h}_j^a} = \frac{1}{m} \cdot \frac{\partial \cos(\bar{\mathbf{h}}^a, \bar{\mathbf{h}}^p)}{\partial \bar{\mathbf{h}}^a}
\label{eq:grad-mean}
\end{equation}
The $1/m$ factor means every token receives the same gradient magnitude. The gradient is dense and uniform (no token is preferentially updated). The model has no mechanism to selectively strengthen discriminative tokens: a function word, a punctuation mark, and a content word all receive the same gradient signal.
 
\paragraph{$\mathrm{MaxSim}$: sparse, selective gradient.}
Under late interaction (Equation~\ref{eq:maxsim}), the gradient with respect to anchor token $j$ is:
\begin{equation}
\footnotesize
\frac{\partial s_{\text{LI}}}{\partial \mathbf{h}_j^a} = \sum_{i=1}^{m_p} \mathbf{1}\!\bigl[j = \operatorname*{argmax}_{j'} \cos(\mathbf{h}_{j'}^a, \mathbf{h}_i^p)\bigr] \cdot \frac{\partial \cos}{\partial \mathbf{h}_j^a}
\label{eq:grad-maxsim}
\end{equation}
Only the tokens selected via $\operatorname*{argmax}$ receive a gradient. Most tokens are not updated at all. The encoder learns which tokens carry discriminative signal because only those tokens participate in the backward pass.
 
\paragraph{PLI: intermediate density.}
Under PLI with patch size $p$ (Equation~\ref{eq:pli}), the gradient combines both regimes:
\begin{equation}
\frac{\partial s_{\text{PLI}}}{\partial \mathbf{h}_j^a} = \frac{1}{p} \cdot \mathbf{1}\!\bigl[\text{patch}(j) \in \operatorname{argmax}\bigr] \cdot \frac{\partial \cos}{\partial \mathbf{h}_j^a}
\label{eq:grad-pli}
\end{equation}
Sparse between patches (only selected patches get gradient), dense within patches (each of the $p$ tokens in a selected patch gets $1/p$).
 
\subsection{The consolidation bottleneck}
\label{sec:theory-bottleneck}
 
Mean pooling's dense gradient creates what we call a consolidation bottleneck. The scoring function only accesses the mean of all tokens. For the encoder to produce a score that distinguishes same-author from different-author passages, it must arrange the hidden states so that their mean already points in a direction that encodes authorship. The encoder must coordinate information across the entire sequence, compressing authorship-relevant features into a form that survives averaging. This compression must happen at some intermediate layer, which we call the \emph{consolidation layer}.
 
$\mathrm{MaxSim}$ has no such bottleneck. The scoring function accesses individual token representations directly, so the encoder can keep refining per-token features through the upper layers without needing to consolidate them into a single direction. The upper layers of a transformer encode more abstract, context-dependent features~\citep{tenney_bert_2019}, so the ability to defer consolidation gives $\mathrm{MaxSim}$ access to richer representations.
 
If our analysis is correct, mean pooling should show a recovery inflection at an earlier layer than $\mathrm{MaxSim}$ when we perform causal patching. Patching below the consolidation layer should destroys the signal (the representation has not yet been compressed). Patching above it should preserve the signal (consolidation is complete). $\mathrm{MaxSim}$ should show a later inflection because there is no pressure to consolidate early.
 
\subsection{Why mean pooling loses information}
\label{sec:theory-info}
 
We can observe mean pooling through an information theory lens and explain why it has less capacity to encode authorship. Mean pooling maps the $m \times d$ token matrix $\mathbf{H}$ to a $d$-dimensional vector $\bar{\mathbf{h}}$. By the data processing inequality, any function of the mean has at most as much mutual information with the author identity $Y$ as a function of the full token matrix:
\begin{equation}
I(Y; \bar{\mathbf{h}}) \leq I(Y; \mathbf{H})
\label{eq:dpi}
\end{equation}
The information loss is strictly positive whenever $\bar{\mathbf{h}}$ is not a sufficient statistic for $Y$. For instance, two passages with identical function-word frequencies but different function-word orderings are indistinguishable under mean pooling (which is permutation-invariant) but distinguishable under  $\mathrm{MaxSim}$ (which preserves positional structure). The information loss is therefore not only theoretical.

\begin{table}[h]
\centering
\small
\begin{tabular}{lcc}
\toprule
\textbf{Model} & $\alpha$ ($\downarrow$) & $u$ ($\downarrow$) \\
\midrule
Pretrained (no fine-tuning) & 0.069 & $-0.171$ \\
Mean pooling                & 0.154 & $-2.000$ \\
LI                          & 0.091 & $-0.331$ \\
PLI $n$-gram 2              & 0.201 & $-0.717$ \\
\bottomrule
\end{tabular}
\caption{Alignment $\alpha$ and uniformity $u$ per model, from \citet{kulumba_halvest_2025}. Lower is better for both.}
\label{tab:au-values}
\end{table}

This capacity gap is reflected in the alignment--uniformity tradeoff (Table~\ref{tab:au-values}). Mean pooling achieves the best uniformity because averaging naturally spreads representations. But it achieves the weakest alignment because it destroys the fine-grained signal needed to cluster same-author passages tightly. LI achieves the tightest alignment because token-level comparison preserves discriminative detail, but the weakest uniformity because the sparse gradient does not prevent representation collapse as aggressively.

%% file: latex/4_setup.tex
\section{Experimental Setup}
\label{sec:setup}

We design a controlled analysis that isolates the scoring mechanism: every model shares one backbone, one corpus, and one loss, differing only in how they turn token representations into a scalar similarity.

\subsection{Models}
\label{sec:setup-models}
 
Every model shares a ModernBERT-base backbone~\citep{warner_smarter_2025} with 23 transformer layers, 149M parameters, and a hidden size of 768. Unless stated otherwise, we use the base-4 split of HALvest-Contrastive~\citep{kulumba_halvest_2025}, a scholarly corpus in which the anchor and positive are drawn from different papers by the same author-set, and the negative is mined from within the same disciplinary field. This design ensures that topical similarity does not confound authorship signal: the model cannot rely on vocabulary overlap to distinguish positives from negatives.
 
\textbf{Layerwise} uses layerwise attention pooling followed by mean pooling and cosine scoring. We use layerwise attention in addition to mean pooling to match the state of the art~\citep{kantharuban_idiolex_2026}. In prior work, layerwise attention adds only a marginal performance gain over raw mean pooling, indicating that the learned layer weights do not overcome the single-vector bottleneck analyzed in \secref{sec:theory-bottleneck}. The gradient with respect to each token still passes through the mean, so the $1/m$ uniform-gradient analysis applies up to a layer-dependent reweighting factor. \textbf{LI} uses token-level $\mathrm{MaxSim}$ with punctuation and padding masked. \textbf{PLI $n{=}2$} uses bigram patch-level $\mathrm{MaxSim}$. \textbf{E5 zero-shot} \citep{wang_text_2024} is included as a control model picked off the shelf. E5 was trained for retrieval, and to a greater extent semantic matching, yielding decorrelated similarity scores from models trained for AA~\citep{kulumba_halvest_2025, kantharuban_idiolex_2026}.

\begin{table}[h]
\centering
\scriptsize
\begin{tabular}{lrrrr}
\toprule
\textbf{Model} & \textbf{R@20} & \textbf{R@100} & \textbf{nDCG@20} & \textbf{nDCG@100} \\
\midrule
Mean pooling       & 0.121 & 0.294 & 0.063 & 0.101 \\
LI              & 0.485 & 0.678 & 0.364 & 0.408 \\
PLI $n{=}2$     & 0.497 & 0.700 & 0.365 & 0.411 \\
E5 (zero-shot)  & 0.167 & 0.269 & 0.124 & 0.146 \\
\bottomrule
\end{tabular}
\caption{HALvest-Contrastive base-4 retrieval performance from \citet{kulumba_halvest_2025}. E5 was tested using mean pooling.}
\label{tab:model-performance}
\end{table}

Table~\ref{tab:model-performance} summarizes retrieval performance. The four-fold Recall@20 gap between mean pooling and LI is the empirical observation we aim to study.
 
\subsection{Probe set construction}
\label{sec:setup-probes}

\begin{figure}[h]
\centering
\includegraphics[width=\columnwidth]{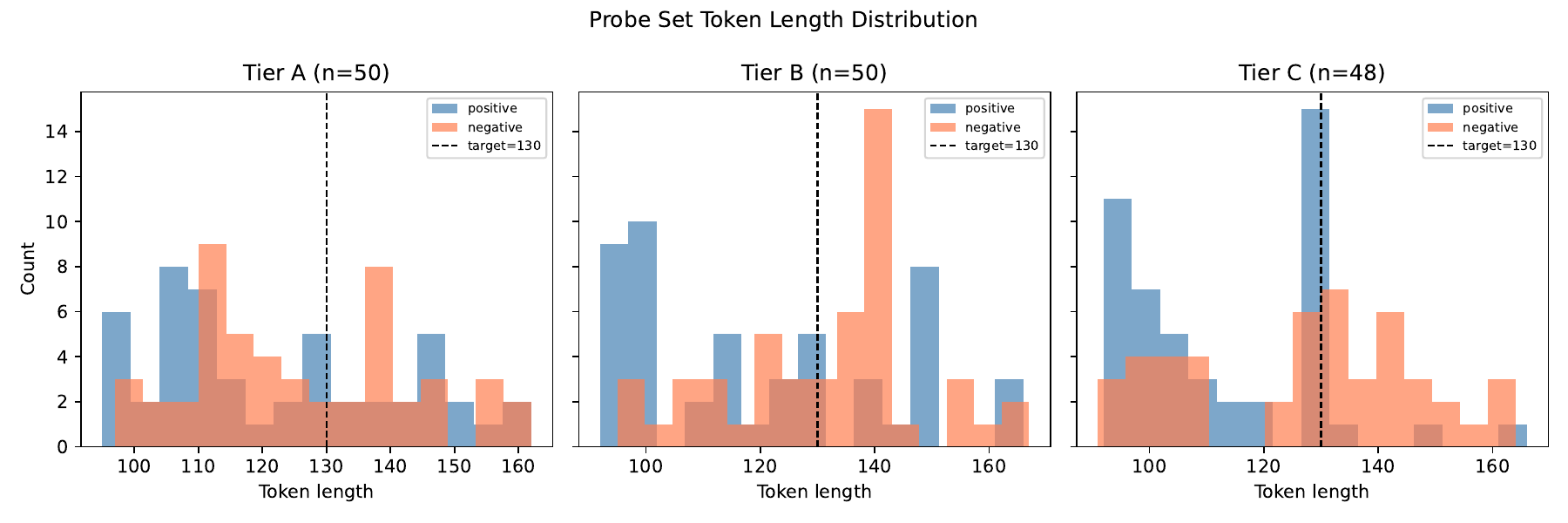}
\caption{Token length distributions for positive (blue) and negative (orange) passages across the three tiers. All passages cluster around the 130-token target.}
\label{fig:length-dist}
\end{figure}

We use a small, controlled set of 148 triplets, not on the full retrieval benchmark to conduct our analysis. Using a curated probe set rather than the full test set allows us to control for confounds (passage length, domain overlap). Triplets are drawn from HALvest-Contrastive base-4 validation, from the ten most frequent author-sets that have at least four distinct documents. Passages target a fixed token length of 130 tokens (Figure~\ref{fig:length-dist}), the positive and negative within each triplet are constrained to differ by at most five tokens after tokenization. Triplets are stratified into three tiers that vary the relationship between the anchor and the negative:

\begin{itemize}
    \item \textbf{Tier~A} ($n{=}50$): the anchor and positive share the same author-set. The negative is written by a completely disjoint author-set from the same scholarly domain. This is the baseline: the model must rely on stylistic signal to distinguish the positive from a topically similar negative written by entirely different authors.
    \item \textbf{Tier~B} ($n{=}50$): the anchor and positive share the same author-set. The negative is written by a partially overlapping author-set that shares at least one author with the anchor's team but is not identical to it. The shared author contributes stylistic signal to both passages, creating a confound. This tier tests whether the model can distinguish full author-set matches from partial ones.
    \item \textbf{Tier~C} ($n{=}48$): the anchor and positive share the same author-set but come from different scholarly domains (anchor in domain $D_1$, positive in domain $D_2$). The negative is written by a disjoint author-set from the anchor's domain~$D_1$. This tests cross-domain authorship recognition: can the model identify the same authors when the vocabulary and conventions shift between disciplines?
\end{itemize}

\begin{table}[h]
\centering
\small
\begin{tabular}{lcccccc}
\toprule
& \multicolumn{2}{c}{\textbf{Tier A}} & \multicolumn{2}{c}{\textbf{Tier B}} & \multicolumn{2}{c}{\textbf{Tier C}} \\
\cmidrule(lr){2-3} \cmidrule(lr){4-5} \cmidrule(lr){6-7}
\textbf{Model} & Fail & $n_+$ & Fail & $n_+$ & Fail & $n_+$ \\
\midrule
Layerwise   & 10\%  & 45 & 44\%  & 28 & 12.5\% & 42 \\
LI          &  2\%  & 49 & 36\%  & 32 &  4.2\% & 46 \\
PLI $n{=}2$ &  2\%  & 49 & 34\%  & 33 &  4.2\% & 46 \\
\bottomrule
\end{tabular}
\caption{Failure rates and effective sample sizes ($n_+$ = correctly-ranked triplets used for patching) per tier. Tier~B is the hardest due to the shared-author confound. Interaction models are more robust than mean pooling across all tiers.}
\label{tab:failure-rates}
\end{table}

Residual patching is only applied to triplets that are correctly ranked (those where the clean model scores the positive above the negative). The effective sample sizes therefore vary by tier and model (Table~\ref{tab:failure-rates}).
 
\subsection{Analyses}
\label{sec:setup-analyses}
 
We apply four analyses to all three fine-tuned models.

\begin{enumerate}
    \item \textbf{LISA probes} train linear classifiers on a separate 10{,}000-passage corpus evaluated on a 2{,}000-passage held-out set, measuring feature availability at each of the 23 layers.
    \item \textbf{Residual stream patching} measures the causal contribution of each layer via rank recovery (Equation~\ref{eq:rank-recovery}) across the 148 probe-set triplets.
    \item \textbf{Score sensitivity} computes the average absolute score change $\overline{|s_{\text{patched}}^{(\ell)} - s_{\text{corrupt}}|}$ per layer, a raw measure of how much the scoring function's output responds to restoring a single layer.
    \item \textbf{Training dynamics} apply patching to eight checkpoints per model (steps 0, 500, 1500, 3000, 5000, 10000, 20000, and final) to track how the depth profile develops during training. It isolates what contrastive fine-tuning adds.
\end{enumerate}

%% file: latex/5_results.tex
\section{Results}
\label{sec:results}

Probing, causal patching, score sensitivity, and training dynamics point to the same conclusion: the performance gap does not arise from what the encoder learns, but from where and how the scorer reads it out.

\subsection{Feature availability is invariant across models}
\label{sec:results-probes}
 
\begin{figure*}[t]
\centering
\begin{subfigure}[b]{0.32\textwidth}
    \includegraphics[width=\textwidth]{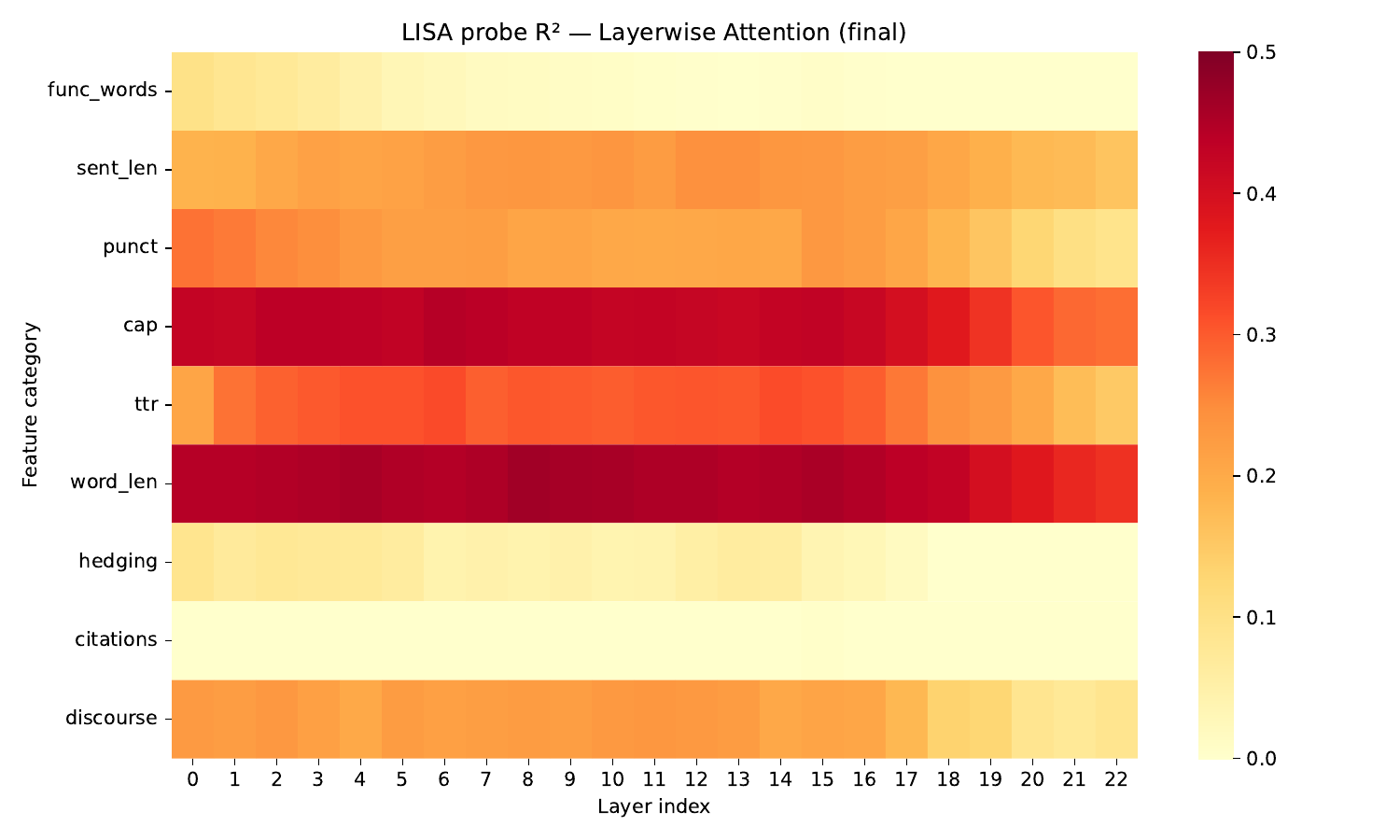}
    \caption{Layerwise (mean pooling)}
    \label{fig:lisa-layerwise}
\end{subfigure}
\hfill
\begin{subfigure}[b]{0.32\textwidth}
    \includegraphics[width=\textwidth]{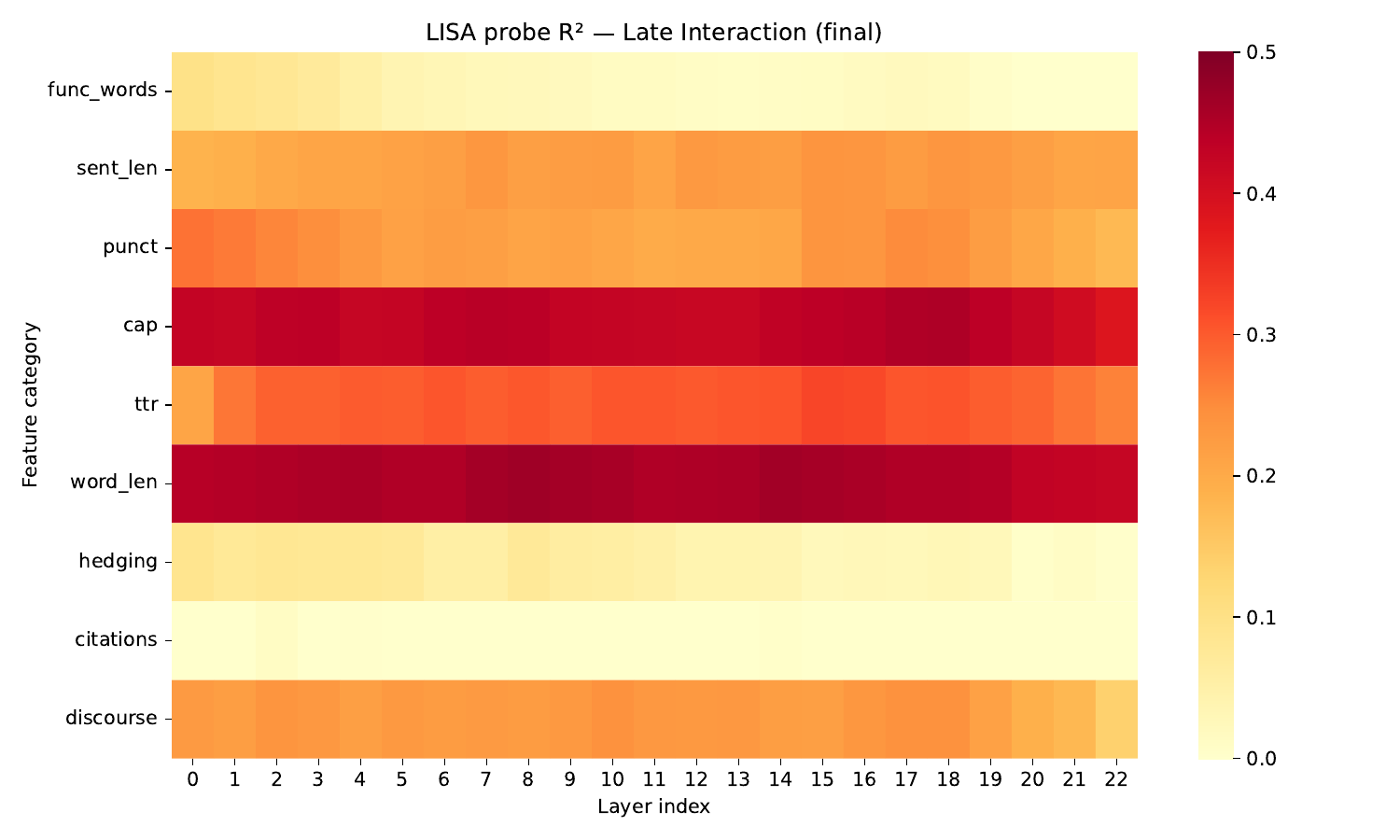}
    \caption{Late Interaction}
    \label{fig:lisa-li}
\end{subfigure}
\hfill
\begin{subfigure}[b]{0.32\textwidth}
    \includegraphics[width=\textwidth]{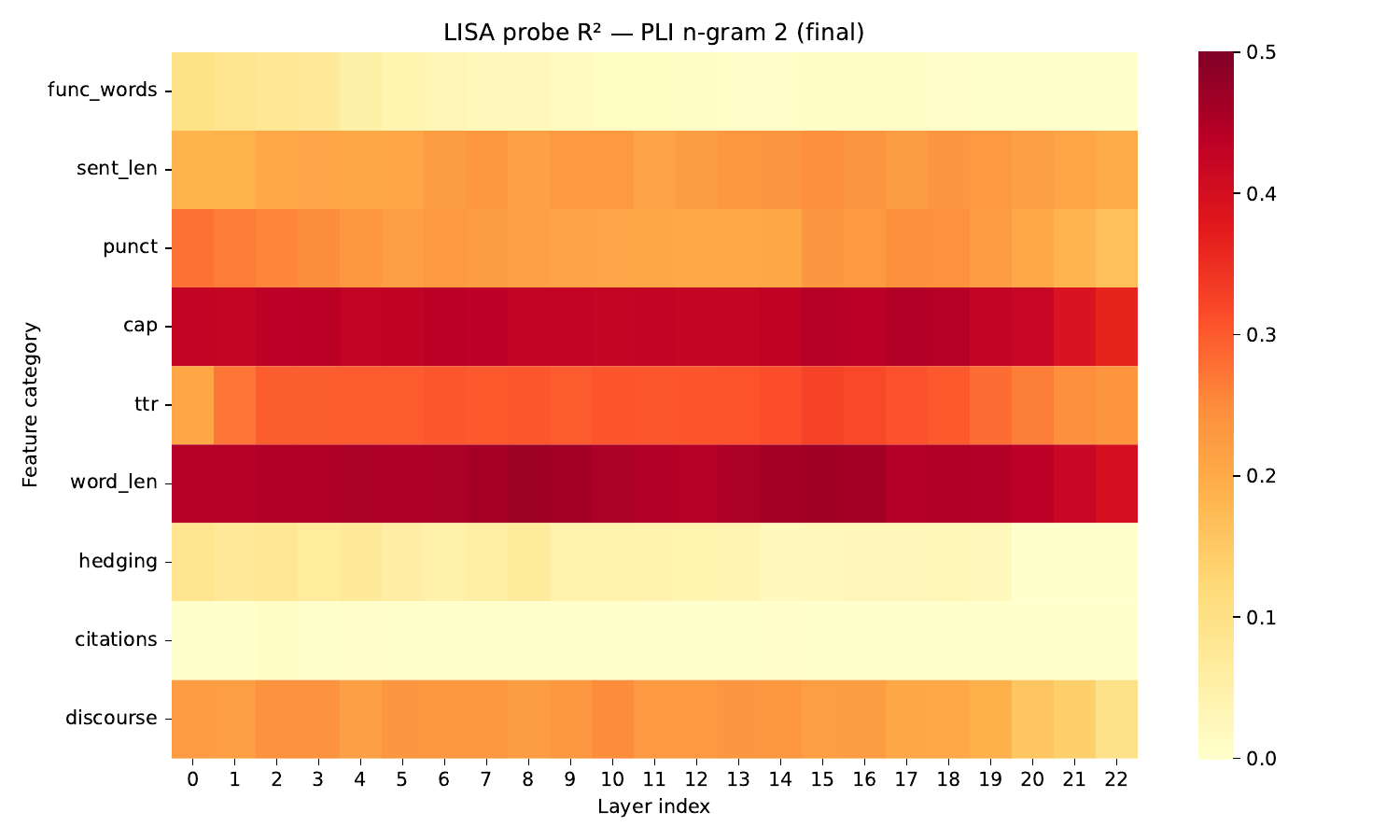}
    \caption{PLI $n{=}2$}
    \label{fig:lisa-pli}
\end{subfigure}
\caption{\textbf{LISA probe $R^2$ heatmaps at the final checkpoint.} Rows are stylistic feature categories. Columns are encoder layers. The three fine-tuned models produce nearly identical heatmaps. Word length is the most readable feature ($R^2 \approx 0.57$), followed by capitalization rate, type--token ratio, and punctuation density.}
\label{fig:lisa-heatmaps}
\end{figure*}

We begin with the question of availability. If the four-fold performance gap between mean pooling and LI arises because LI causes the encoder to learn better stylistic representations, then the LISA probes should show higher $R^2$ for LI than for mean pooling, at least at some layers. It is, however, not the case. Figure~\ref{fig:lisa-heatmaps} shows the probe heatmaps for all three fine-tuned models. The heatmaps are visually indistinguishable. The top features, word length, capitalization, type--token ratio, punctuation density, and function-word frequency, achieve the same $R^2$ at the same layers across all models. The E5 control produces a similar pattern (Appendix~\ref{sec:app:top-lisa}. Thus, the pretrained backbones already makes these stylistic features linearly readable under our probes.
 
This weakens the first hypothesis from the introduction. Within the stylistic features and linear probes used here, we do not find evidence that different scorers lead to different recoverable representations. The pretrained ModernBERT backbone already encodes these features and contrastive fine-tuning does not create them, regardless of the scoring function. The four-fold performance gap is therefore more plausibly explained by differences in how these features are used than by differences in what the encoder learned.
 
\subsection{Causal patching reveals a scoring-dependent depth profile}
\label{sec:results-patching}
 
\begin{figure*}[t]
\centering
\begin{subfigure}[b]{0.32\textwidth}
    \includegraphics[width=\textwidth]{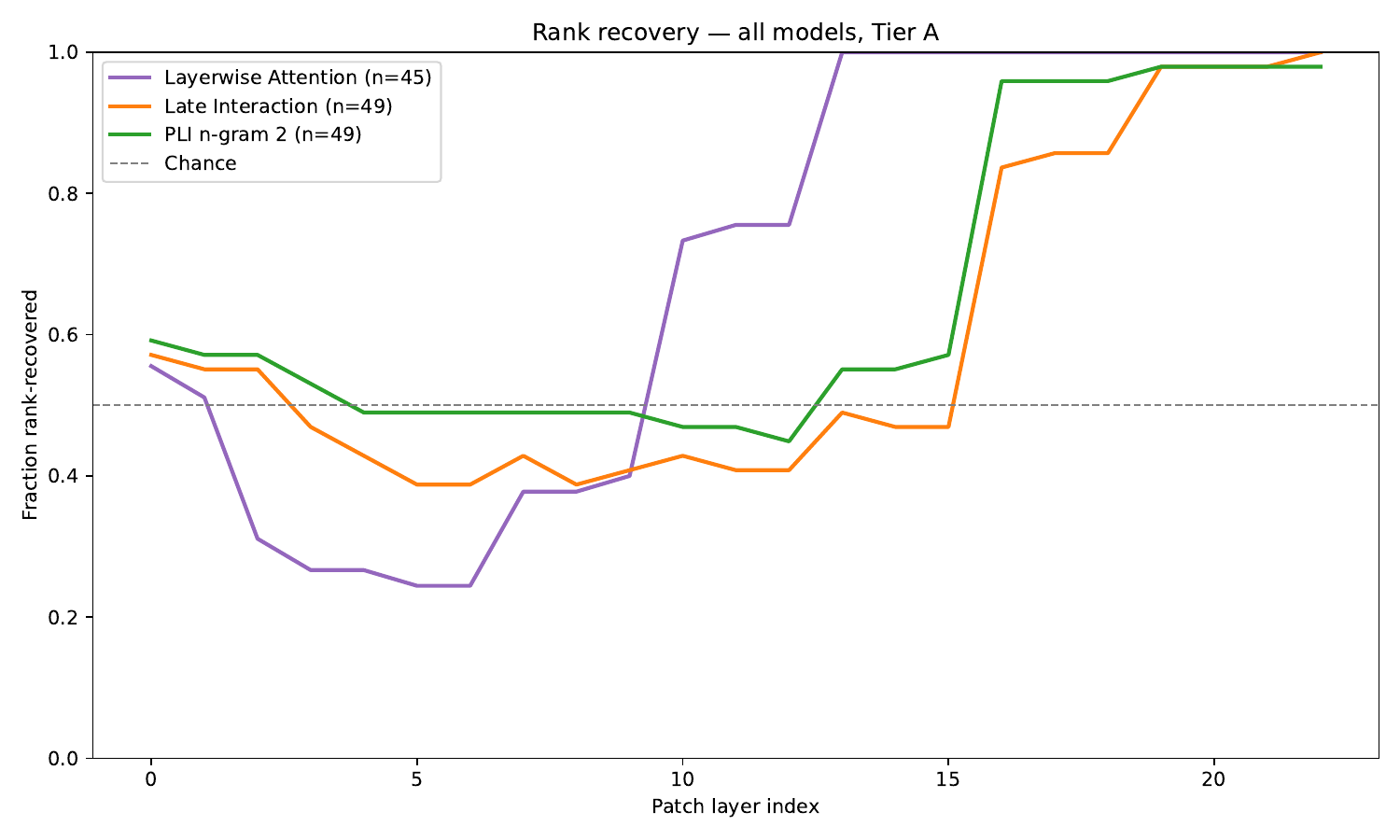}
    \caption{Tier~A (same domain)}
    \label{fig:rank-all-a}
\end{subfigure}
\hfill
\begin{subfigure}[b]{0.32\textwidth}
    \includegraphics[width=\textwidth]{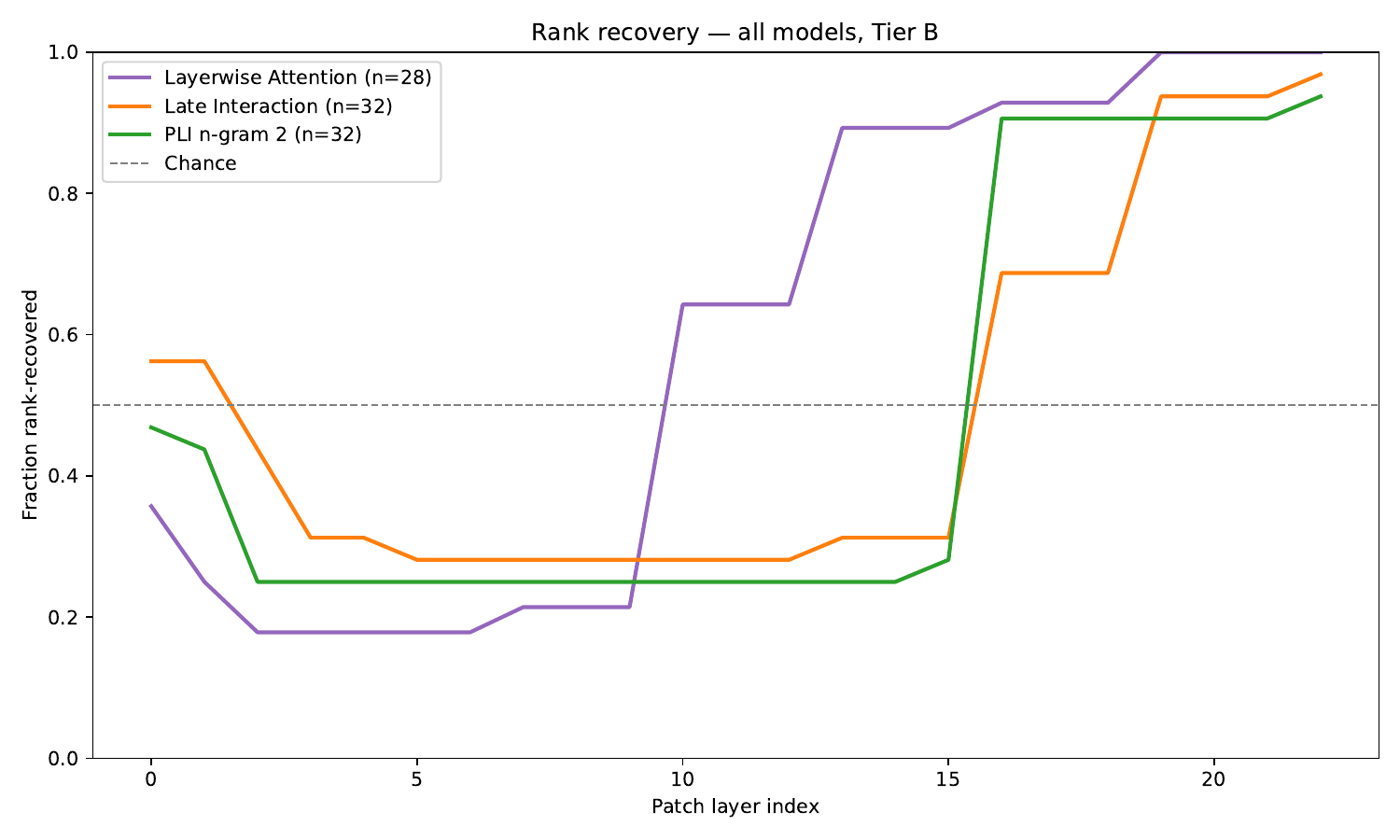}
    \caption{Tier~B (shared-author confound)}
    \label{fig:rank-all-b}
\end{subfigure}
\hfill
\begin{subfigure}[b]{0.32\textwidth}
    \includegraphics[width=\textwidth]{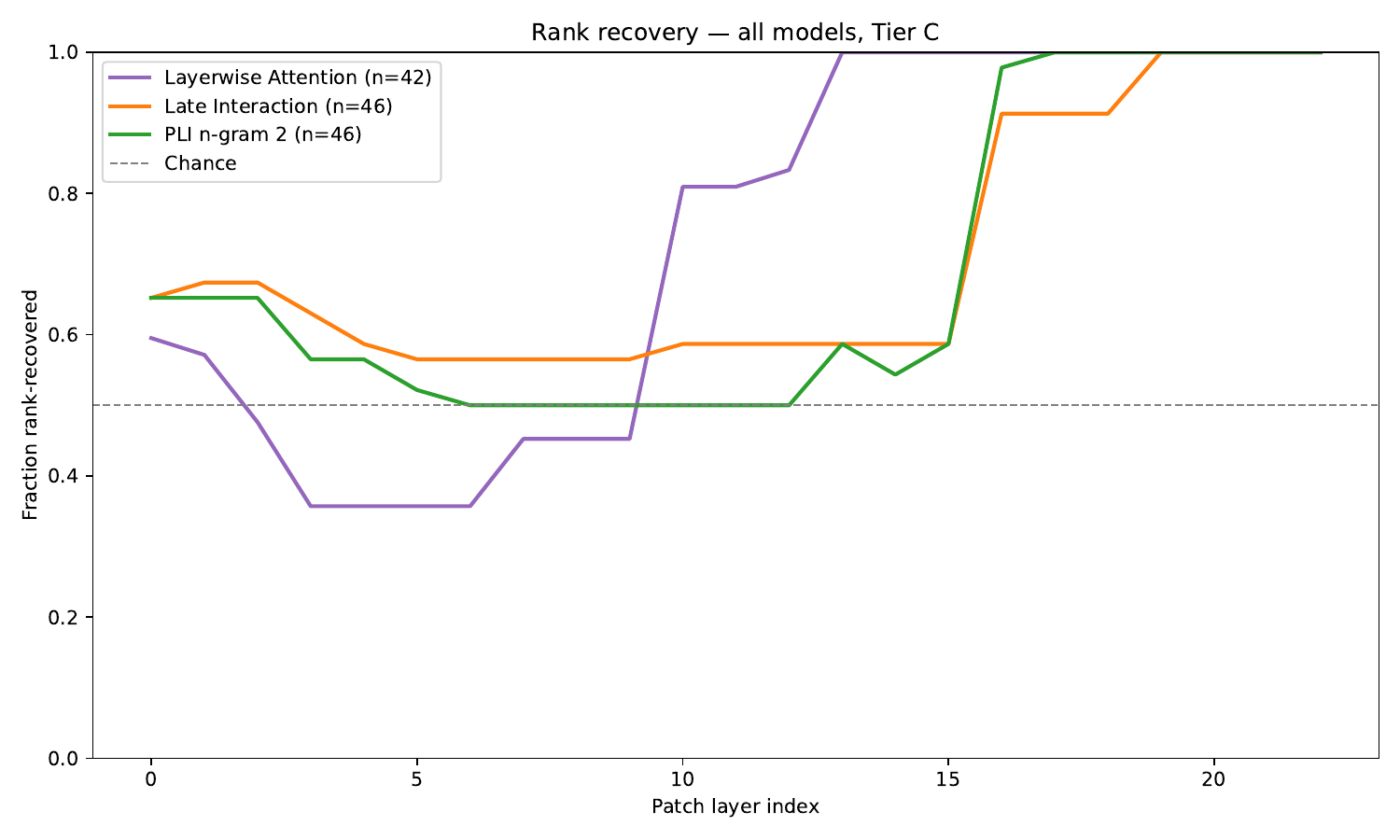}
    \caption{Tier~C (cross-domain)}
    \label{fig:rank-all-c}
\end{subfigure}
\caption{\textbf{Rank recovery across the three models.} Each panel shows one tier. Purple: layerwise (mean pooling), orange: LI, green: PLI $n{=}2$. Dashed line: chance (0.5). Mean pooling crosses chance at layer~9, while both interaction models cross at layers~14--16. The six-layer gap is consistent across all three tiers.}
\label{fig:rank-all}
\end{figure*}

\paragraph{Layerwise (mean pooling)} follows an S-shape. The curve crosses random guess at approximately layer~9 and reaches near-perfect recovery by layer~13. This pattern is consistent across all three tiers. On Tier~C, all models show slightly above-chance performance at the very first layers (0–-2). This is consistent with early layers encoding shallow syntactic statistics~\citep{jawahar_what_2019} that carry distributional authorship signal even when domain-specific vocabulary shifts. In Tiers A and B, topical overlap between anchor and negative may mask this early signal.
 
\paragraph{Late interaction} shows a qualitatively similar S-curve but with a later inflection. Rank recovery stays below random guess until approximately layer~15, then steeply rises to $\geq 0.90$ by layer~20. The below-chance dip at layers~3--12 is deeper than for layerwise (recovery~$\approx 0.3$--$0.4$): corrupting these layers actively misleads the token-level $\mathrm{MaxSim}$ scoring.
 
\paragraph{PLI $n{=}2$} tracks LI closely. The inflection falls at layers~14--16, effectively indistinguishable from LI given the sample size.
 
\paragraph{We define the consolidation point} as the earliest layer at which rank recovery exceeds 0.75. By this criterion, mean pooling consolidates at layer~10, while LI and PLI consolidate at layers~16 and 15 respectively. This is consistent with the prediction from \secref{sec:theory-bottleneck}: dense, uniform gradients favor earlier consolidation while sparse, selective gradients allow later consolidation. PLI $n{=}2$ does not interpolate between the two, it falls squarely in the interaction regime, consistent with the patch $\operatorname{argmax}$'s selection dominating the intra-patch averaging (Equation~\ref{eq:grad-pli}).

\subsection{Score sensitivity confirms two regimes}
\label{sec:results-sensitivity}

\begin{figure*}[t]
\centering
\begin{subfigure}[b]{0.32\textwidth}
    \includegraphics[width=\textwidth]{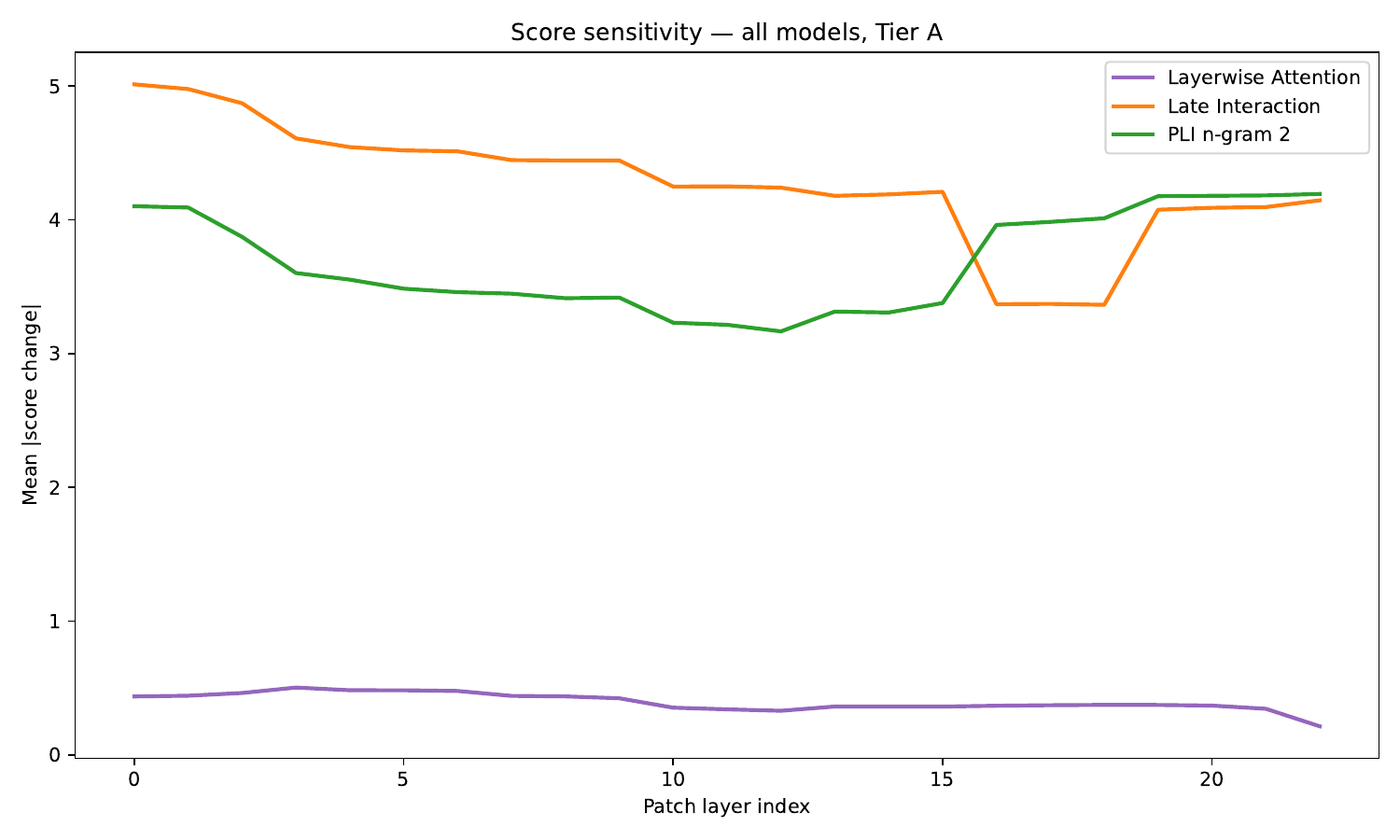}
    \caption{Tier~A}
\end{subfigure}
\hfill
\begin{subfigure}[b]{0.32\textwidth}
    \includegraphics[width=\textwidth]{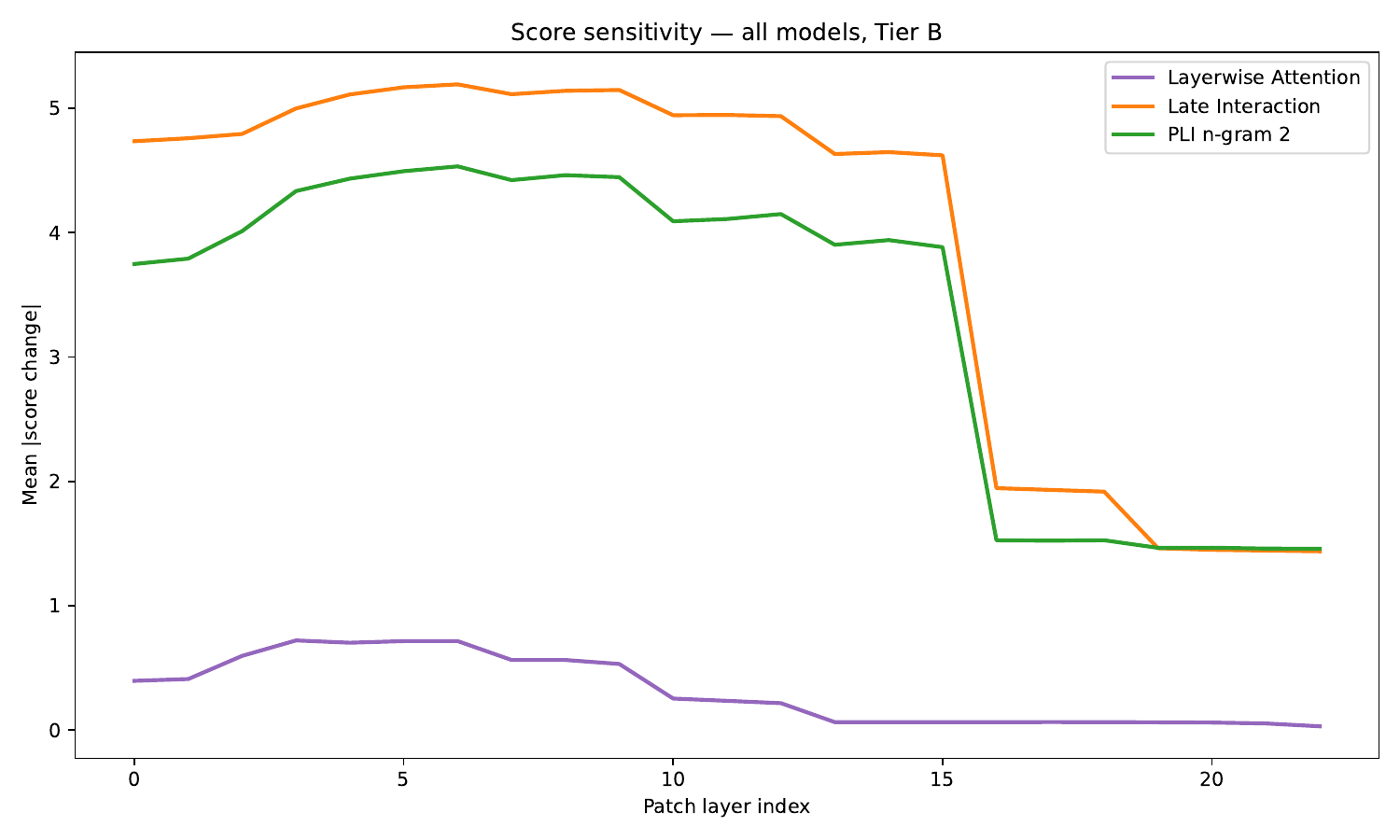}
    \caption{Tier~B}
\end{subfigure}
\hfill
\begin{subfigure}[b]{0.32\textwidth}
    \includegraphics[width=\textwidth]{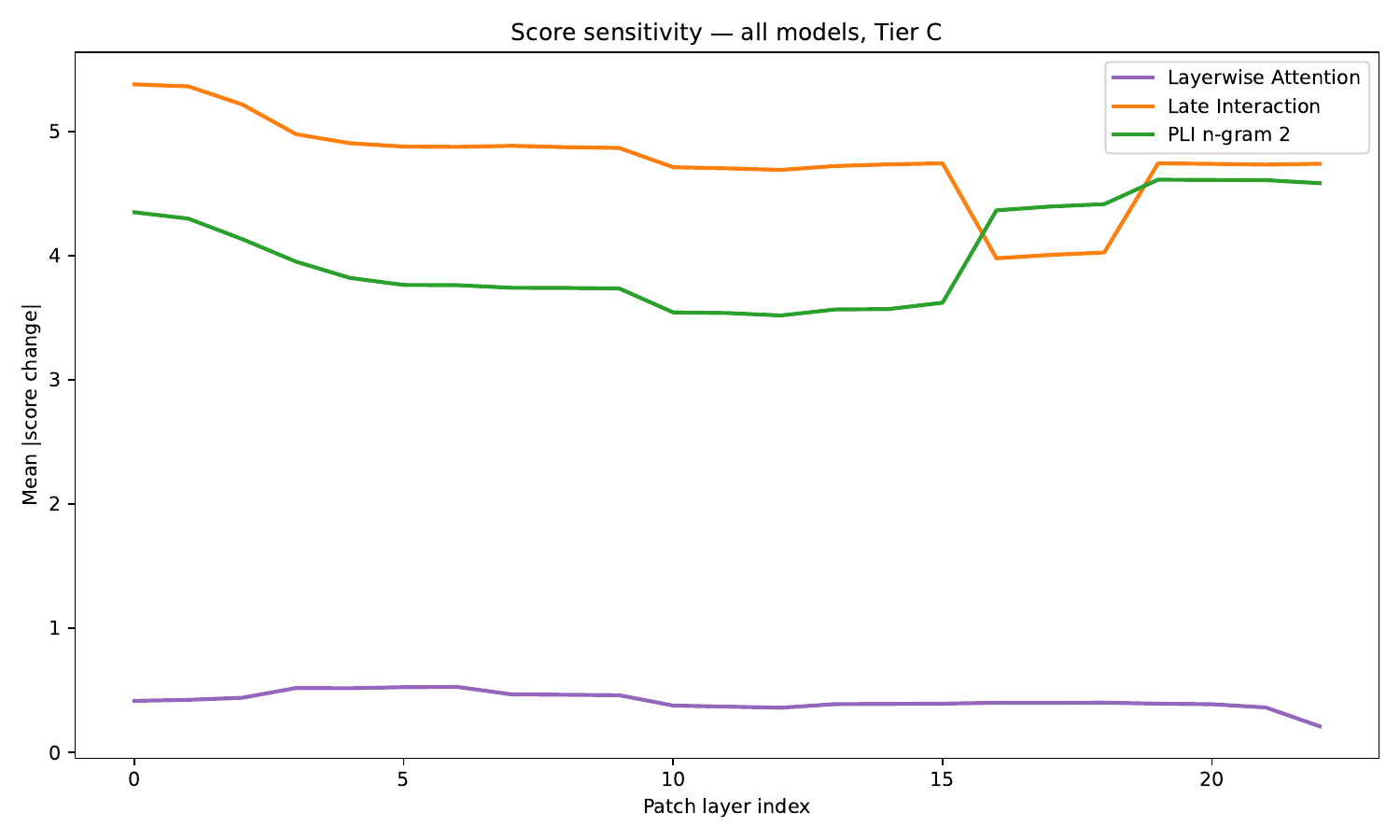}
    \caption{Tier~C}
\end{subfigure}
\caption{\textbf{Score sensitivity per layer.} Mean $|s_{\text{patched}}^{(\ell)} - s_{\text{corrupt}}|$ when restoring clean activations at layer $\ell$. LI (orange) is most sensitive, PLI (green) is intermediate, layerwise (purple) is an order of magnitude lower.}
\label{fig:sensitivity}
\end{figure*}
 
Score sensitivity provides a complementary view: rather than asking whether patching recovers the correct ranking, it asks how much the score changes in absolute terms (Figure~\ref{fig:sensitivity}). The ordering is consistent across all tiers: LI is most sensitive, PLI is intermediate and layerwise is an order of magnitude lower.Mean pooling compresses representations so heavily that restoring a single layer barely moves the mean. $\mathrm{MaxSim}$ reads individual tokens, so a layer-level perturbation can change which tokens are selected by the $\operatorname{argmax}$, producing a large score shift. PLI sits 10–-20\% below LI, consistent with intra-patch averaging partially smoothing perturbations before the patch-level $\operatorname{argmax}$.

\subsection{Training dynamics reveal three learning trajectories}
\label{sec:results-dynamics}
 
\begin{figure*}[t]
\centering
\begin{subfigure}[b]{\textwidth}
    \includegraphics[width=\textwidth]{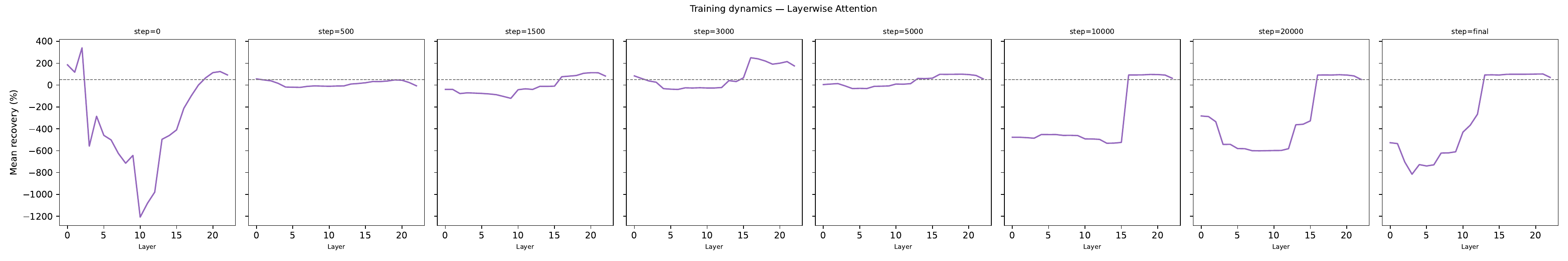}
    \caption{Layerwise (mean pooling): top-down monotonic. Upper layers learn first, the inflection migrates downward during training.}
    \label{fig:dynamics-layerwise}
\end{subfigure}
\vspace{0.3em}
\begin{subfigure}[b]{\textwidth}
    \includegraphics[width=\textwidth]{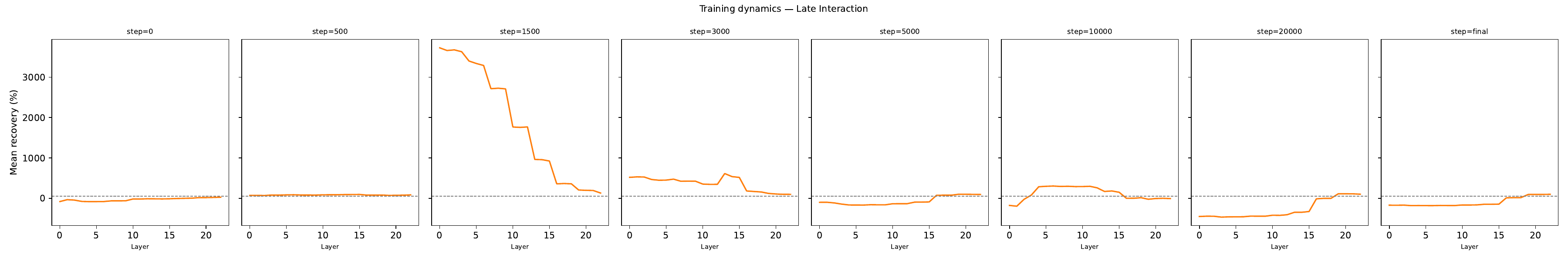}
    \caption{Late Interaction: transient early-layer spike at step~1500, then signal migration to upper layers. The trajectory is consistent with an initial reliance on shallow lexical matches, followed by a shift toward deeper representations.}
    \label{fig:dynamics-li}
\end{subfigure}
\vspace{0.3em}
\begin{subfigure}[b]{\textwidth}
    \includegraphics[width=\textwidth]{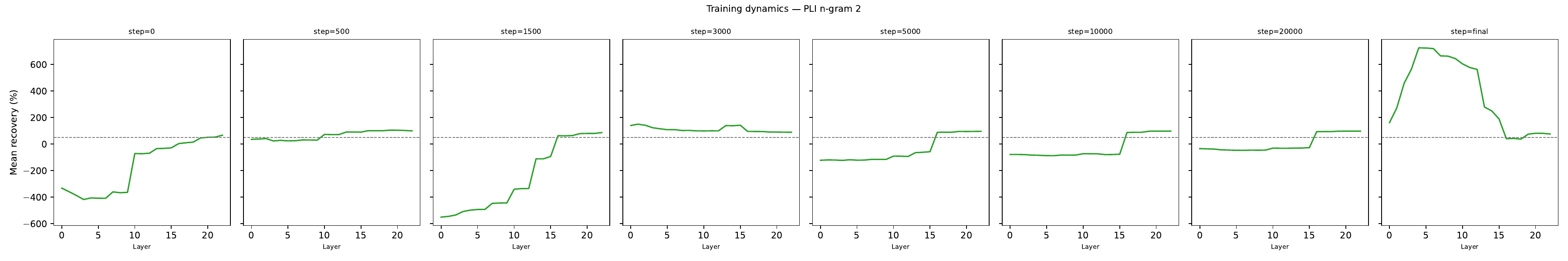}
    \caption{PLI $n{=}2$: gradual emergence, no transient spike. The final checkpoint shows a distinctive mid-layer hump (layers~10--15) absent in both other models.}
    \label{fig:dynamics-pli}
\end{subfigure}
\caption{\textbf{Training dynamics.} Mean percentage recovery across Tier~A triplets at eight checkpoints. Each subplot is one checkpoint. x-axis: layer index; y-axis: mean recovery. Percentage recovery is used here because rank recovery is binary and too coarse to track gradual signal emergence at early checkpoints. The y-axis extremes reflect the known instability of percentage recovery (\secref{sec:bg-recovery}).}
\label{fig:dynamics}
\end{figure*}
 
The patching analysis so far shows the final-checkpoint depth profile. To understand how that profile develops, we apply the same analysis to intermediate checkpoints (Figure~\ref{fig:dynamics}).

\paragraph{ Mean pooling (Figure~\ref{fig:dynamics-layerwise})} learns top-down. At step~500, recovery is concentrated at the uppermost layers. As training progresses, the inflection migrates downward: layer~15 by step~3000, layer~13 by step~10{,}000, layer~9 at the final checkpoint. The model progressively recruits deeper layers to consolidate earlier, consistent with the consolidation bottleneck. The dense gradient initially refines the layers closest to the scoring function, then gradually shapes earlier layers.
 
\paragraph{Late interaction (Figure~\ref{fig:dynamics-li})} shows a distinctive behavior. At step~1500, recovery spikes at layers~5--10. This suggests the model initially exploits shallow lexical matches: $\mathrm{MaxSim}$ can propagate gradient through exact token matches at negligible cost, providing a cheap authorship signal from lower layers. As hard negatives increase in difficulty during training, this shortcut becomes insufficient, and the model shifts to deeper, more contextualized representations. By step~5000, this transient behavior is suppressed and recovery concentrates at layers~19+. The model learns to defer to deeper, more abstract representations, abandoning the shallow shortcut.
 
\paragraph{PLI $n{=}2$ (Figure~\ref{fig:dynamics-pli}).} Bigram-patch $\mathrm{MaxSim}$ shows a third pattern with no early spike: the intra-patch averaging smooths out the shallow matches that LI exploits. Recovery emerges gradually at the upper layers. The final checkpoint shows a mid-layer hump (layers~10--15) unique to PLI, possibly reflecting the two-level structure of its gradient (Equation~\ref{eq:grad-pli}). Mid-layer patch representations carry authorship signal that neither the token-level first moment (mean pooling) nor the individual tokens ($\mathrm{MaxSim}$) would use.

%% file: latex/6_related_work.tex
\section{Related Work}
\label{sec:related}

\paragraph{Authorship attribution.}
Neural AA has evolved from classification~\citep{burrows_delta_2002, schler_effects_2006} to contrastive learning~\citep{wegmann_same_2022, kantharuban_idiolex_2026, huertas-tato_isolating_2024}, with increasing focus on topic confounding~\citep{wegmann_does_2021, rivera-soto_learning_2021}. Our work is not the first attempt of the AA community at interpretability~\citep{alshomary_layered_2025, alshomary_latent_2025}, but is, to the best of our knowledge the first one to use mechanistic interpretability tools and gradient analysis to derive performance and training behavior from encoder models.

\paragraph{Probing versus causal analysis.}
Linear probes~\citep{belinkov_probing_2022} are widely used to study what information neural representations encode, but the link between probe accuracy and actual model behavior is contested~\citep{hewitt_designing_2019, ravichander_probing_2021}. Activation patching~\citep{vig_investigating_2020, meng_locating_2022, wang_interpretability_2022} provides a causal alternative: it asks whether information is necessary, not merely decodable. Our \emph{availability} against \emph{use} dissociation contributes to this debate by showing that all probed features are equally available across models with very different task performance.

%% file: latex/7_discussion.tex
\section{Discussion}
\label{sec:discussion}

The availability–use dissociation reframes AA as an information readout problem. In this setup, the pretrained encoder already makes the stylistic features linearly readable. What differs is whether the scoring function can access them at the right depth and with enough capacity.

\paragraph{Availability against use.}
In this setting, probing accuracy is not a sufficient proxy for task performance when models differ in their scoring mechanism. All four models, three fine-tuned and one off-the-shelf control, produce nearly identical probe heatmaps (Figure~\ref{fig:lisa-heatmaps}) while differing dramatically in retrieval performance (Table~\ref{tab:model-performance}). The main question is not, therefore, which model encodes more stylistic information, but which scoring mechanism can effectively read it out. Path patching at the attention-head level~\citep{goldowsky-dill_localizing_2023} could further localize how stylistic signal flows through the encoder, though this addresses a finer-grained question than the one considered here.
 
\paragraph{Why interaction beats pooling.}
The gradient analysis (\secref{sec:theory-gradient}) and the information-theoretic argument (\secref{sec:theory-info}) converge: mean pooling discards higher-order structure by compressing to a single vector, while $\mathrm{MaxSim}$ preserves token-level granularity. The causal depth profiles confirm this: consolidation at layer 9 versus layers 15–16. The probe results (Figure~\ref{fig:lisa-heatmaps}), patching curves (Figure~\ref{fig:rank-all}), score sensitivity analysis (Figure~\ref{fig:sensitivity}), and training dynamics (Figure~\ref{fig:dynamics}),all converge on the same explanation.
 
\paragraph{PLI in the interaction regime.}
PLI $n=2$ falls in the same causal regime as LI, with nearly identical recovery inflections. This suggests that the patch-level $\operatorname{argmax}$ dominates the effect of local averaging inside each patch. The alignment and uniformity results (Table~\ref{tab:au-values}) are also consistent with this interpretation, since PLI remains much closer to LI than to mean pooling in embedding-space geometry. Whether larger patches shift consolidation earlier remains open, though the theory predicts that they should gradually approach the pooling regime.
 
Overall, the results suggest that the main bottleneck in contrastive authorship attribution is not whether stylistic information exists in the encoder, but whether the scoring mechanism can preserve and exploit it.